\documentclass{article}
\usepackage[utf8]{inputenc}

\title{Arbitrated Indirect Treatment Comparisons}
\author{Yixin Fang\thanks{Correspondence: 1 North Waukegan Rd, North Chicago, IL 60064; {email: yixin.fang@abbvie.com}} \quad and \quad Weili He\\
Data and Statistical Sciences, AbbVie Inc.}
\date{\today}

\usepackage{amsmath}
\usepackage{natbib}
\usepackage{graphicx}
\usepackage{amssymb}
\usepackage{setspace}
\usepackage{booktabs}
\usepackage{multirow}
\usepackage{bm}

\usepackage[a4paper, margin=1.2in]{geometry}

\DeclareMathAlphabet\mathbfcal{OMS}{cmsy}{b}{n}

\begin{document}

\doublespacing

\maketitle

\begin{abstract}

Matching-adjusted indirect comparison (MAIC) has been increasingly employed in health technology assessments (HTA). By reweighting subjects from a trial with individual participant data (IPD) to match the covariate summary statistics of another trial with only aggregate data (AgD), MAIC facilitates the estimation of a treatment effect defined with respect to the AgD trial population. This manuscript introduces a new class of methods, termed \emph{arbitrated indirect treatment comparisons}, designed to address the ``MAIC paradox''---a phenomenon highlighted by Jiang et al.~(2025). The MAIC paradox arises when different sponsors, analyzing the same data, arrive at conflicting conclusions regarding which treatment is more effective. The underlying issue is that each sponsor implicitly targets a different population. To resolve this inconsistency, the proposed methods focus on estimating treatment effects in a common target population, specifically chosen to be the \emph{overlap population}.

\end{abstract}


\section{Introduction}

In support of health technology assessment (HTA) submissions, we often need to conduct indirect treatment comparisons (ITC). One common type of ITC is population-adjusted indirect comparisons (PAIC), in which individual patient data (IPD) in one trial and aggregate data (AgD) in the other trial are used to adjust for differences between trials in the distribution of variables that influence outcome \citep{phillippo2016nice}. The two most popular PAIC methods are the matching adjusted indirect comparison (MAIC) and the simulated treatment comparison (STC) \citep{caro2010no, signorovitch2010comparative, ishak2015simulation, phillippo2016nice}.

Two scenarios in which PAIC can be applied have been identified, resulting in anchored PAIC and unanchored PAIC \citep{signorovitch2012matching}. In the anchored PAIC scenario, a randomized controlled clinical trial (RCT) with IPD comparing treatment A with comparator C (the AC trial) is utilized, along with another RCT with AgD comparing treatment B with comparator C (the BC trial). In the unanchored PAIC scenario, a single-arm trial with IPD (which may come from the treatment arm of an RCT) is used together with another single-arm trial with AgD (which may originate from the treatment arm of another RCT). In this paper, the focus is placed on anchored PAIC, and it is noted that all discussions are readily extendable to unanchored PAIC.

The principal challenge faced in the conduct of PAIC is the presence of imbalances in effect modifiers between two trials, namely the AC trial and the BC trial. Accordingly, MAIC and STC have been developed to facilitate the adjustment for these imbalances in effect modifiers between the two trials. Nevertheless, a paradox, referred to as the ``MAIC paradox,'' has been demonstrated by \cite{jiang2025critical}, who stated the following: 

\begin{quote} ``However, when there are imbalances in effect modifiers with different magnitudes of modification across treatments, contradictory conclusions may arise if MAIC is performed with the IPD and AgD swapped between trials.''
\end{quote}

The root cause of the MAIC paradox is considered to be the following:
\begin{itemize}
\item When MAIC is conducted by sponsor A using the IPD of the AC trial and the AgD of the BC trial, the targeted population is defined by the BC trial;
\item When MAIC is conducted by sponsor B using the IPD of the BC trial and the AgD of the AC trial, the targeted population is defined by the AC trial.
\end{itemize}

In general, apart from differences in the distribution of effect modifiers across populations, there may be additional underlying factors that lead to varying magnitudes of treatment effects in different populations. For example, the principle of transportability may be violated, implying that results obtained from one population may not necessarily generalize to another \citep{pearl2009causality}.

Therefore, after the above root cause has been identified, a practical solution to the paradox is provided by the consideration of a common population for conducting MAIC by both sponsors simultaneously. Evidently, the overlap population between the two trials can be regarded as an appropriate common population. A class of methods to estimate the average treatment effect for the overlap population (ATO) will be proposed in this paper. 

However, in order for the proposed methods to be implemented, the involvement of a third party, referred to as the arbitrator (for example, an HTA body), is required. The arbitrator is responsible for ensuring that MAIC is conducted by both sponsors, targeting at a common population (for example, the overlap population of the AC trial population and the BC trial population). It should be noted that some methods necessitate the sharing of IPD data with the arbitrator, while other methods do not require such sharing.

The demonstrative example provided by Jiang \emph{et al.}~(2025) is revisited, with the  hypothetical dataset presented in Table \ref{Tab:data}, in which race (Black vs.~non-Black) is considered the sole effect modifier. The distributions of race in the AC trial and in the BC trial are given by
\begin{align*}
\mathbb{P}_{\mbox{\tiny AC}}(\mbox{Black})=\frac{1}{3},\quad \mathbb{P}_{\mbox{\tiny AC}}(\mbox{non-Black})=\frac{2}{3};\\
\mathbb{P}_{\mbox{\tiny BC}}(\mbox{Black})=\frac{2}{3},\quad \mathbb{P}_{\mbox{\tiny BC}}(\mbox{non-Black})=\frac{1}{3}.
\end{align*}

\begin{table}[ht]\caption{The hypothetical dataset from Jiang \emph{et al.}~(2025)}\label{Tab:data}
\medskip
\centering
\renewcommand{\arraystretch}{1}
\begin{tabular}{l l r r r r r r }
\toprule
 & &  \multicolumn{3}{c}{\textbf{AC Trial}} & \multicolumn{3}{c}{\textbf{BC Trial}}  \\
\cmidrule(lr){3-5} \cmidrule(lr){6-8} 
 & & Drug A & Drug C & logOR & Drug B & Drug C & logOR \\
\midrule
 \multirow{5}{*}{Black} & $Y = 0$ & 180 & 80 & & 240 & 160 &  \\
 & $Y = 1$ & 20 & 120 & & 160 & 240 &  \\
 & $n$ & 200 & 200 & & 400 & 400 & \\
\midrule
\multirow{5}{*}{non-Black} & $Y = 0$ & 80 & 40 & & 100 & 20 &  \\
 & $Y = 1$ & 320 & 360 & & 100 & 180 &  \\
 & $n$ & 400 & 400 & & 200 & 200 &  \\
\midrule
Overall & Survival rate & 43.3\% & 20\% & 1.12 & 56.7\% & 30\% & 1.12 \\
\bottomrule
\end{tabular}
\end{table}

When sponsor A conducts MAIC, the weights for black patients and for the non-black patients in the AC trial are calculated as 
\begin{align*}
    w_{\mbox{\tiny AC}}(\mbox{Black})=\frac{1}{600},\quad w_{\mbox{\tiny AC}}(\mbox{non-Black})=\frac{1}{2400}.
\end{align*}
Under these weights, the weighted distribution of race in the AC trial is 
\begin{align*}
    \mathbb{P}^w_{\mbox{\tiny AC}}(\mbox{Black})=\frac{400(1/600)}{800(1/2400)+400(1/600)}=\frac{2}{3},\quad \mathbb{P}^w_{\mbox{\tiny AC}}(\mbox{non-Black})=\frac{1}{3},
\end{align*}
which matches the distribution of race in the BC trial. Under these weights, the weighted estimate of the treatment effect for drug A vs.~drug C (in terms of the logarithm of the odds ratio of the survival rate) in the AC trial is
\begin{align*}
    \log\left[\frac{\left(180(1/600)+80(1/2400)\right)\left(120(1/600)+360(1/2400)\right)}{\left(20(1/600)+320(1/2400)\right)\left(80(1/600)+40(1/2400)\right)}\right]=1.54.
\end{align*}
The MAIC estimate of the treatment effect for drug A vs.~drug B is then equal to 
\begin{align*}
    \widehat{\theta}_{\mbox{\tiny AB}\mid \mbox{\tiny AC}}=1.54-1.12=0.42.
\end{align*}
This indicates that drug A is superior to drug B in improving the odds of survival when sponsor A performs MAIC.

On the other hand, when sponsor B conducts MAIC, the weights for black patients and for the non-black patients in the BC trial are calculated as 
\begin{align*}
     w_{\mbox{\tiny BC}}(\mbox{Black})=\frac{1}{2400},\quad
    w_{\mbox{\tiny BC}}(\mbox{non-Black})=\frac{1}{600}.
\end{align*}
Under these weights, the weighted distribution of race in the BC trial is 
\begin{align*}
    \mathbb{P}^w_{\mbox{\tiny BC}}(\mbox{Black})=\frac{800(1/2400)}{800(1/2400)+400(1/600)}=\frac{1}{3},\quad \mathbb{P}^w_{\mbox{\tiny BC}}(\mbox{non-Black})=\frac{2}{3},
\end{align*}
which matches the distribution of race in the AC trial. Under these weights, the weighted estimate of the treatment effect for drug B vs.~drug C in the BC trial is
\begin{align*}
\log\left[\frac{\left(240(1/2400)+100(1/600)\right)\left(240(1/2400)+180(1/600)\right)}{\left(160(1/2400)+100(1/600)\right)\left(160(1/2400)+20(1/600)\right)}\right]=1.52.
\end{align*}
The MAIC estimate of the treatment effect for drug A vs.~drug B is then equal to 
\begin{align*}
    \widehat{\theta}_{\mbox{\tiny BA}\mid {\mbox{\tiny BC}}}=1.52-1.12=0.40.
\end{align*}
This indicates that drug B is superior to drug A in improving the odds of survival when sponsor B performs MAIC. 

Therefore, using the same method of MAIC and analyzing the same data, sponsor A demonstrates that drug A is superior to drug B, whereas sponsor B demonstrates that drug B is superior to drug A. This is the MAIC paradox \citep{jiang2025critical}.

To address the MAIC paradox, suppose that the arbitrator has access to IPD from both the AC and BC trials. The arbitrator would then consider the overlap population, as shown in Table~\ref{Tab:data-overlap}. For this target population, the arbitrator estimates the treatment effect in terms of log-odds ratio:
\begin{align*}
\widehat{\theta}_{\mbox{\tiny AB}} = 1.30 - 1.30 = 0.
\end{align*}
Therefore, with IPD available from both trials, the arbitrated ITC reveals a treatment effect of zero comparing drugs A and B for the overlap population, indicating no difference between the two drugs within the overlap population.

\begin{table}[ht]\caption{The overlap population from the hypothetical dataset}\label{Tab:data-overlap}
\medskip
\centering
\renewcommand{\arraystretch}{1}
\begin{tabular}{l l r r r r r r }
\toprule
 & &  \multicolumn{3}{c}{\textbf{AC Trial}} & \multicolumn{3}{c}{\textbf{BC Trial}}  \\
\cmidrule(lr){3-5} \cmidrule(lr){6-8} 
 & & Drug A & Drug C & logOR & Drug B & Drug C & logOR \\
\midrule
 \multirow{5}{*}{Black} & $Y = 0$ & 180 & 80 & & 120 & 80 &  \\
 & $Y = 1$ & 20 & 120 & & 80 & 120 &  \\
 & $n$ & 200 & 200 & & 200 & 200 & \\
\midrule
\multirow{5}{*}{non-Black} & $Y = 0$ & 40 & 20 & & 100 & 20 &  \\
 & $Y = 1$ & 160 & 180 & & 100 & 180 &  \\
 & $n$ & 200 & 200 & & 200 & 200 &  \\
\midrule
Overall & Survival rate & 55\% & 25\% & 1.30 & 55\% & 25\% & 1.30 \\
\bottomrule
\end{tabular}
\end{table}

However, in practice, it is often infeasible for the arbitrator to obtain IPD from both the AC and BC trials. Consequently, there is a need for methods that enable arbitrated ITC without access to IPD from both studies.

The remainder of the paper is organized as follows. Section 2 introduces the concept of overlap weights. Section 3 proposes two arbitrated MAIC methods. Section 4 revisits the hypothetical example and presents a simulation study. Section 5 provides a brief discussion. 

\section{Overlap Weights} 

\subsection{Overlap Weights for Two Treatment Groups}

\cite{li2018balancing} proposed a general class of weights---the balancing weights---which are designed to balance the weighted distributions of the covariates between two treatment groups. An important implication of using the balancing weights is that they define a general class of estimands. 

Consider a sample of $n$ patients, each assigned to one of two treatment groups, for which covariate-balanced comparisons are of interest. Let $Z_i=z$ denote a binary variable indicating the group membership of subject $i$, where $z=1$ or $z=0$. For each patient, an outcome $Y_i$ and a set of covariates $\bm{X}_i=(X_{i1}, \dots, X_{ip})^\top\in \mathbb{R}^p$ are observed. The propensity score \citep{rosenbaum1983central} is the probability of assignment to the treatment group given the covariates, 
\begin{align*}
    e(\bm{x})=\mathbb{P}(Z_i=1\mid \bm{X}_i=\bm{x}).
\end{align*}

Let $f(\bm{x})$ be the marginal density of the covariates $\bm{X}$ with respect to a base measure $\mu$---a product of counting measures for categorical covariates and Lebesgue measure for continuous variables. For any pre-specified function $h(\bm{x})$, the density of the form 
\begin{align*}
    f_h^*(\bm{x})=\frac{f(\bm{x})h(\bm{x})}{\int f(\bm{x})h(\bm{x})\mu(d\bm{x})} 
\end{align*}
defines the target population. With respect to this target population, the estimand is defined as 
\begin{align*}
    \theta_h&=\int [\mathbb{E}(Y\mid Z=1, \bm{X}=\bm{x})-\mathbb{E}(Y\mid Z=0, \bm{X}=\bm{x})]f_h^*(\bm{x})\mu(d\bm{x}).
\end{align*}

Here the discussion on the identifiability assumptions (consistency, exchangeability, and positivity) using the language of potential outcomes is omitted \citep{imbens2015causal, hernan2020causal, fang2024causal}, in order to focus on the estimation. 

Some examples of $h(\bm{x})$, which lead to some commonly considered estimands, are presented. When $h(\bm{x})=1$, the estimand $\theta_h$ is the average treatment effect (ATE). When $h(\bm{x})=e(\bm{x})$, the estimand $\theta_h$ is the average treatment effect among the treated (ATT). When $h(\bm{x})=1-e(\bm{x})$, the estimand $\theta_h$ is the average treatment effect among the controls (ATC). When $h(\bm{x})=e(\bm{x})(1-e(\bm{x}))$, the estimand $\theta_h$ is the average treatment effect for the overlap population (ATO). 

The estimation of ATO will be explored in more details, because it is relevant to the arbitrated ITC methods to be proposed later. Let $f_z(\bm{x})=\mathbb{P}(\bm{X}=\bm{x}\mid Z=z)$ be the density of $\bm{X}$ in the $Z=z$ group, $z=1, 0$. Then, by the Bayesian formula, we have
\begin{align*}
    f_1(\bm{x})\propto f(\bm{x})e(\bm{x})\quad \mbox{and} \quad f_0(\bm{x})\propto f(\bm{x})(1-e(\bm{x})).
\end{align*}
The overlap weights \citep{li2018balancing} are defined as
\begin{align*}
    w_1(\bm{x})=1-e(\bm{x}) \quad \mbox{and} \quad w_0(\bm{x})=e(\bm{x}),
\end{align*}
for the $Z=1$ group and the $Z=0$ group, respectively. 

Under the overlap weights, the weighted distributions of $\bm{X}$ in the two groups are 
\begin{align*}
    f_1(\bm{x})w_1(\bm{x})&\propto f(\bm{x})e(\bm{x})\cdot (1-e(\bm{x}))=f(\bm{x})\cdot e(\bm{x})(1-e(\bm{x})),\\
     f_0(\bm{x})w_0(\bm{x})&\propto f(\bm{x})(1-e(\bm{x}))\cdot e(\bm{x})=f(\bm{x})\cdot e(\bm{x})(1-e(\bm{x})),
\end{align*}
which are equal to the distribution $\bm{X}$ of the same target population, the overlap population. Therefore, the definition of overlap weights implies the choice of $h(\bm{x})=e(\bm{x})(1-e(\bm{x}))$.

By contrast, the inverse propensity score weights are defined as 
\begin{align*}
    w'_1(\bm{x})=\frac{1}{e(\bm{x})} \quad \mbox{and} \quad w'_0(\bm{x})=\frac{1}{1-e(\bm{x})},
\end{align*}
for the $Z=1$ group and the $Z=0$ group, respectively. Under the inverse propensity score weights, the weighted distributions of $\bm{X}$ in the two groups are 
\begin{align*}
    f_1(\bm{x})w'_1(\bm{x})&\propto f(\bm{x})e(\bm{x})\cdot \frac{1}{e(\bm{x})}=f(\bm{x})\cdot 1,\\
     f_0(\bm{x})w_0(\bm{x})&\propto f(\bm{x})(1-e(\bm{x}))\cdot \frac{1}{1-e(\bm{x})}=f(\bm{x})\cdot 1,
\end{align*}
which are both equal to the marginal distribution of $\bm{X}$ over the combined population. Therefore, the definition of inverse propensity score weights implies the choice of $h(\bm{x})=1$.

\begin{figure}[h]
\centering
\caption{Illustration of four target populations in a two-arm study}
\label{fig_1}
\includegraphics[width=0.9\linewidth]{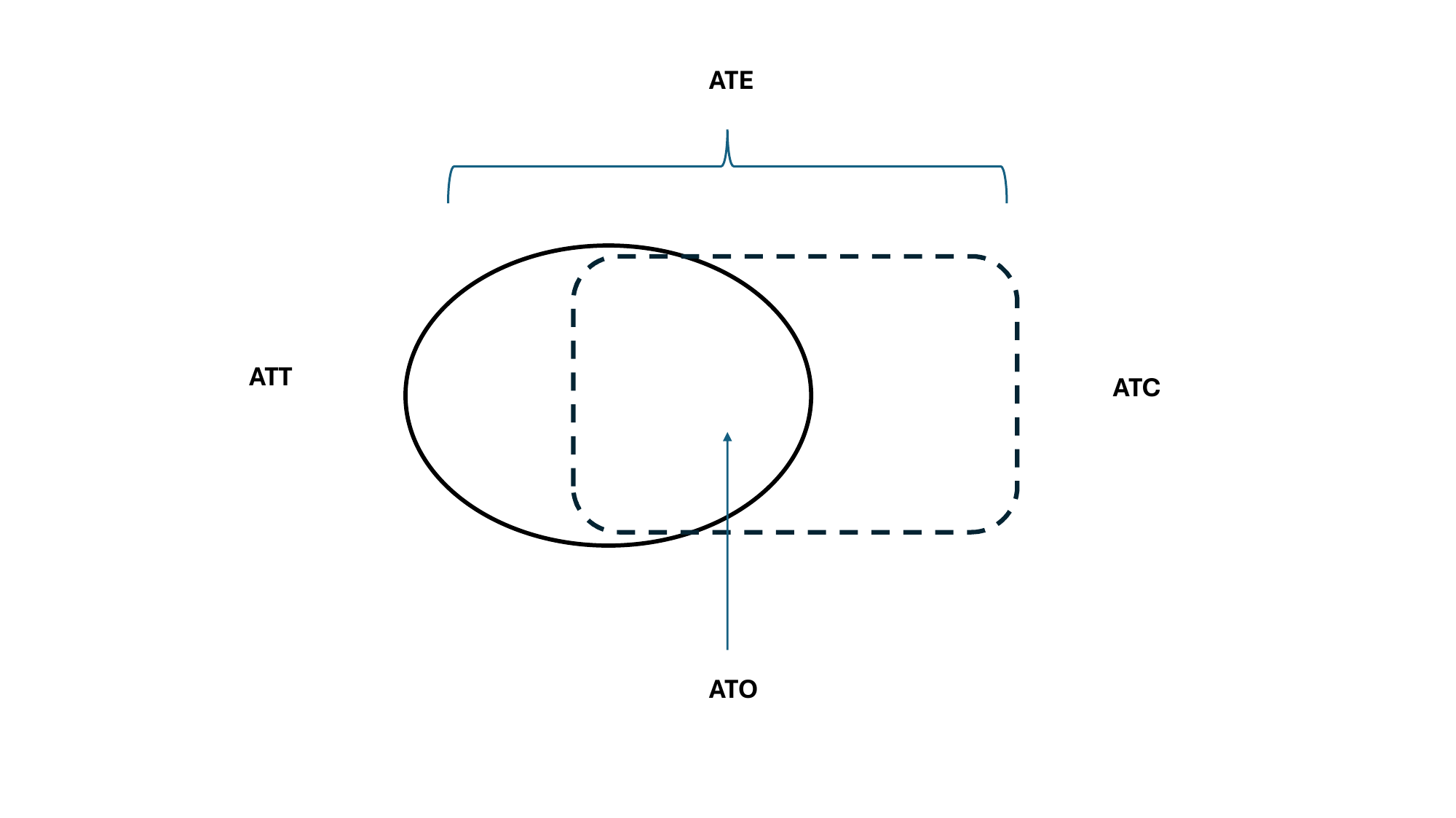}
\end{figure}

Figure \ref{fig_1} illustrates four target populations, the population of $Z=1$, the population of $Z=0$, the combined population, and the overlap population, based on which the associated estimands (ATT, ATC, ATE, and ATO) are defined.

\subsection{Extension of Overlap Weights to Two Trials}

Consider two trials: the AC trial, consisting of $n_{\text{\tiny AC}}$ patients, and the BC trial, consisting of $n_{\text{\tiny BC}}$ patients. Let $T_i = t$ denote a binary variable indicating the trial membership of patient $i$, where $t = 1$ corresponds to the AC trial and $t = 0$ to the BC trial, for $i = 1, \dots, n_{\text{\tiny AC}} + n_{\text{\tiny BC}}$. 

For each patient, the following variables are observed: an outcome $Y_i$, a treatment indicator $Z_i$, and a vector of effect modifiers $\bm{X}_i = (X_{i1}, \dots, X_{ip})^\top\in \mathbb{R}^p$. In the AC trial, $Z_i$ takes values in $\{\text{A}, \text{C}\}$, whereas in the BC trial, $Z_i$ takes values in $\{\text{B}, \text{C}\}$.

The concept of overlap weights for two treatment groups, as discussed previously, can be naturally extended to the scenario involving two clinical trials. Before delving into the detailed description below, please note the following correspondence in advance:
\begin{itemize}
    \item Population $T=1$ vs. Population $T=0$ $\Longleftrightarrow$ Population $Z=1$ vs. Population $Z=0$;
    \item Enrollment propensity score $\varepsilon(\bm{x})$ $\Longleftrightarrow$ Assignment propensity score $e(\bm{x})$;
    \item Overlap weights $(\omega_1(\bm{x}), \omega_0(\bm{x}))$ $\Longleftrightarrow$ Overlap weights $(w_1(\bm{x}), w_0(\bm{x}))$.
\end{itemize}

The probability of trial enrollment given covariates $\bm{X}$ is defined by the following enrollment propensity score:
\begin{align*}
    \varepsilon(\bm{x}) = \mathbb{P}(T_i = 1 \mid \bm{X}_i = \bm{x}).
\end{align*}

Consequently, the overlap weights for each trial are defined as follows:
\begin{align*}
    \omega_1(\bm{x}) = 1 - \varepsilon(\bm{x})\quad \mbox{and}\quad  \omega_0(\bm{x}) = \varepsilon(\bm{x}),
\end{align*}
where $\omega_1(\bm{x})$ and $\omega_0(\bm{x})$ are used for weighting patients in the AC trial ($T=1$) and in the BC trial ($T=0$), respectively.  

Let $\varphi(\bm{x})$ represent the marginal density of $\bm{X}$ in the combined population of both trials. Thus, the densities of $\bm{X}$ for the AC and BC trial are defined as $\varphi_1(\bm{x})$ and $\varphi_0(\bm{x})$, respectively, which have the following properties:
\begin{align*}
    \varphi_1(\bm{x})\propto \varphi(\bm{x})\varepsilon(\bm{x}) \quad \mbox{and}\quad \varphi_0(\bm{x})\propto \varphi(\bm{x})(1-\varepsilon(\bm{x})).
\end{align*}
Therefore, under the overlap weights, the weighted distributions of $\bm{X}$ in the two trials are 
\begin{align*}
    \varphi_1(\bm{x})\omega_1(\bm{x})&\propto \varphi(\bm{x})\varepsilon(\bm{x})\cdot (1-\varepsilon(\bm{x}))=\varphi(\bm{x})\cdot \varepsilon(\bm{x})(1-\varepsilon(\bm{x})),\\
     \varphi_0(\bm{x})\omega_0(\bm{x})&\propto \varphi(\bm{x})(1-\varepsilon(\bm{x}))\cdot \varepsilon(\bm{x})=\varphi(\bm{x})\cdot \varepsilon(\bm{x})(1-\varepsilon(\bm{x})),
\end{align*}
which are equal to the distribution of $\bm{X}$ over the same target population, the overlap population. Therefore, the definition of overlap weights implies the choice of $h(\bm{x})=\varepsilon(\bm{x})(1-\varepsilon(\bm{x}))$.

The estimand of interest is the average treatment effect between drug A and drug B for the overlap population of the AC and BC populations, defined as
\begin{align*}
    \theta^{\text{\tiny AB}}_{\text{\tiny ATO}} 
    = \frac{
        \int \left[ \mathbb{E}(Y \mid Z = \text{A}, \bm{X} = \bm{x}) - \mathbb{E}(Y \mid Z = \text{B}, \bm{X} = \bm{x}) \right] \varphi(\bm{x}) \varepsilon(\bm{x})(1 - \varepsilon(\bm{x})) \, \mu(d\bm{x})
    }{
        \int \varphi(\bm{x}) \varepsilon(\bm{x})(1 - \varepsilon(\bm{x})) \, \mu(d\bm{x})
    },
\end{align*}
where $\varphi(\bm{x})$ represents the marginal density of $\bm{X}$ in the combined population of both trials.

To facilitate indirect comparison, the treatment contrast between A and B can be decomposed using a common comparator C. Accordingly, the estimand can be rewritten as
\begin{align*}
    \theta^{\text{\tiny AB}}_{\text{\tiny ATO}} = 
    \theta^{\text{\tiny AC}}_{\text{\tiny ATO}} - 
    \theta^{\text{\tiny BC}}_{\text{\tiny ATO}},
\end{align*}
where
\begin{align*}
    \theta^{\text{\tiny AC}}_{\text{\tiny ATO}} &= 
    \frac{
        \int \left[ \mathbb{E}(Y \mid Z = \text{A}, \bm{X} = \bm{x}) - \mathbb{E}(Y \mid Z = \text{C}, \bm{X} = \bm{x}) \right] \varphi(\bm{x}) \varepsilon(\bm{x})(1 - \varepsilon(\bm{x})) \, \mu(d\bm{x})
    }{
        \int \varphi(\bm{x}) \varepsilon(\bm{x})(1 - \varepsilon(\bm{x})) \, \mu(d\bm{x})
    }, \\
    \theta^{\text{\tiny BC}}_{\text{\tiny ATO}} &= 
    \frac{
        \int \left[ \mathbb{E}(Y \mid Z = \text{B}, \bm{X} = \bm{x}) - \mathbb{E}(Y \mid Z = \text{C}, \bm{X} = \bm{x}) \right] \varphi(\bm{x}) \varepsilon(\bm{x})(1 - \varepsilon(\bm{x})) \, \mu(d\bm{x})
    }{
        \int \varphi(\bm{x}) \varepsilon(\bm{x})(1 - \varepsilon(\bm{x})) \, \mu(d\bm{x})
    }.
\end{align*}

The above reasoning leads to two key results: based on IPD from the AC trial,  standard weighted estimators using the overlap weights $\omega_1(\bm{x})$ consistently estimate the estimand $\theta^{\text{\tiny AC}}_{\text{\tiny ATO}}$, while based on IPD from the BC trial, standard weighted estimators using the overlap weights $\omega_0(\bm{x})$ consistently estimate the estimand $\theta^{\text{\tiny BC}}_{\text{\tiny ATO}}$. Taken together, these results provide a formal justification---detailed in the Appendix---for the proposed methods discussed in the next section.

By contrast, the inverse propensity score weights are defined as 
\begin{align*}
    \omega'_1(\bm{x})=\frac{1}{\varepsilon(\bm{x})} \quad \mbox{and} \quad \omega'_0(\bm{x})=\frac{1}{1-\varepsilon(\bm{x})},
\end{align*}
for the $T=1$ trial and the $T=0$ trial, respectively. Under the inverse propensity score weights, the weighted distributions of $\bm{X}$ in the two trials are 
\begin{align*}
    \varphi_1(\bm{x})\omega'_1(\bm{x})&\propto \varphi(\bm{x})\varepsilon(\bm{x})\cdot \frac{1}{\varepsilon(\bm{x})}=\varphi(\bm{x})\cdot 1,\\
     \varphi_0(\bm{x})\omega_0(\bm{x})&\propto \varphi(\bm{x})(1-\varepsilon(\bm{x}))\cdot \frac{1}{1-\varepsilon(\bm{x})}=\varphi(\bm{x})\cdot 1,
\end{align*}
which are both equal to the marginal distribution of $\bm{X}$ over the combined population. Therefore, the definition of inverse propensity score weights implies the choice of $h(\bm{x})=1$.

\begin{figure}[!ht]
\centering
\caption{Illustration of four target populations for two clinical trials}
\label{fig_2}
\includegraphics[width=0.9\linewidth]{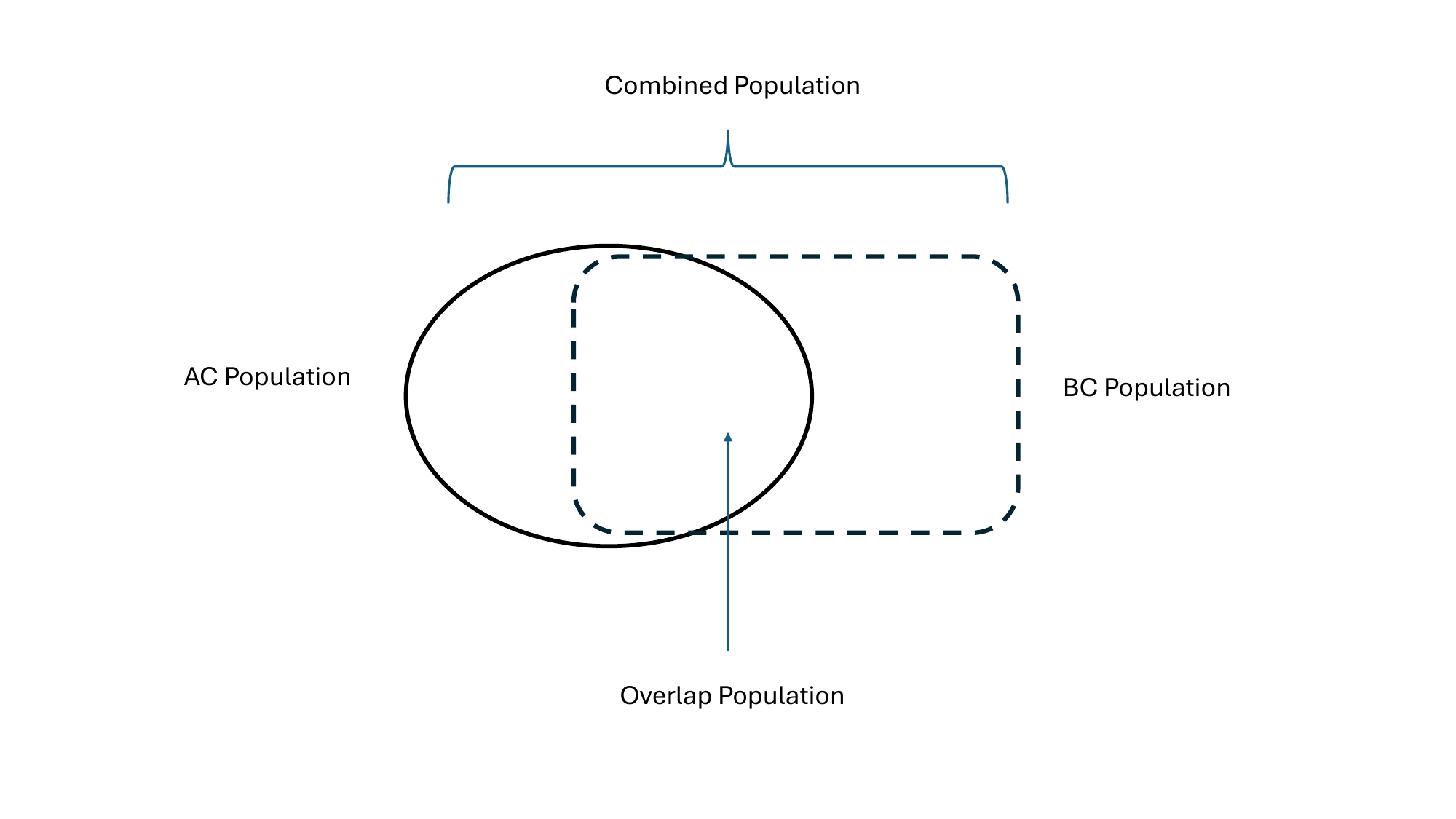}
\end{figure}

Figure \ref{fig_2} illustrates four target populations, the population of $T=1$, the population of $T=0$, the combined population, and the overlap population. From this Figure, both the combined population and the overlap population can serve as a common target population to address the MAIC paradox. However, since the inverse propensity score weights are unstable when the propensity score approaches zero at certain values of $\bm{x}$, we do not recommend using the combined population. Notably, because the overlap weights are confined to the interval $[0,1]$, they remain stable. Therefore, we advocate for the use of the overlap population.

\section{Arbitrated MAIC Methods}

\subsection{Arbitrator Having Access to IPD of Covariates}

Consider the scenario where both sponsors are willing to share their IPD of covariates with the arbitrator. After receiving the IPD of covariates from both sponsors, the arbitrator can run a propensity score model (e.g., logistic regression model) based on the following data:
\begin{align*}
    \left\{(\bm{X}_i, T_i),\quad i=1, \dots, n_{\mbox{\tiny{AC}}}+n_{\mbox{\tiny{BC}}}\right\},
\end{align*}
where $T_i=1$ for $i=1, \dots, n_{\mbox{\tiny{AC}}}$ and $T_i=0$ for $i=n_{\mbox{\tiny{AC}}}+1, \dots, n_{\mbox{\tiny{AC}}}+n_{\mbox{\tiny{BC}}}$. 

After the propensity score is trained, the following fitted propensity scores are obtained: 
\begin{align*}
    \left\{\widehat{\varepsilon}(\bm{X}_i),\quad i=1, \dots, n_{\mbox{\tiny{AC}}}+n_{\mbox{\tiny{BC}}}\right\}.
\end{align*}
Then the arbitrator shares the following overlap weights with the two sponsors separately: 
\begin{align*}
    \boldsymbol{\omega}_1&=\{1-\widehat{\varepsilon}(\bm{X}_i), i=1, \dots, n_{\mbox{\tiny{AC}}}\}\quad \mbox{(with sponsor A)}, \\
    \boldsymbol{\omega}_0&=\{\widehat{\varepsilon}(\bm{X}_i), i=n_{\mbox{\tiny{AC}}}+1, \dots, n_{\mbox{\tiny{AC}}}+n_{\mbox{\tiny{BC}}}\}\quad \mbox{(with sponsor B)}. 
\end{align*}

After receiving the corresponding overlap weights from the arbitrator, the two sponsors proceed to conduct their own MAIC separately. Any existing MAIC method recommended in \cite{phillippo2016nice} is applicable. 

For example, the generalized estimating equations (GEE) \citep{liang1986longitudinal} method can be used to perform a weighted analysis, providing both a point estimator and a robust standard error estimator (a.k.a.~the sandwich estimator). Implementation can be performed using the \texttt{ geeglm} function from the package R \texttt{geepack} \citep{hojsgaard2006r} with a \texttt{weights} option to perform a weighted analysis, or using the SAS \citep{SAS} procedure GENMOD with a WEIGHT statement to perform a weighted analysis.

Thus, sponsor A conducts MAIC using the weights $\boldsymbol{\omega}_1$, producing an estimate $\widehat{\theta}^{\text{\tiny AC}}_{\text{\tiny ATO}}$ and an estimate of its standard error $\widehat{\mbox{SE}}^{\text{\tiny AC}}_{\text{\tiny ATO}}$ for the estimand $\theta^{\text{\tiny AC}}_{\text{\tiny ATO}}$, while sponsor B conducts MAIC using the weights $\boldsymbol{\omega}_0$, producing an estimate $\widehat{\theta}^{\text{\tiny BC}}_{\text{\tiny ATO}}$ and an estimate of its standard error $\widehat{\mbox{SE}}^{\text{\tiny BC}}_{\text{\tiny ATO}}$ for the estimand $\theta^{\text{\tiny BC}}_{\text{\tiny ATO}}$. The sponsors then share their results with the arbitrator. 

Finally, after receiving the results from the sponsors, the arbitrator obtains the following point estimate and an estimate of its standard error, 
\begin{align*}
    \widehat{\theta}^{\text{\tiny AB}}_{\text{\tiny ATO}}&=\widehat{\theta}^{\text{\tiny AC}}_{\text{\tiny ATO}}-\widehat{\theta}^{\text{\tiny BC}}_{\text{\tiny ATO}},\\
    \widehat{\mbox{SE}}^{\text{\tiny AB}}_{\text{\tiny ATO}}&=\sqrt{\left(\widehat{\mbox{SE}}^{\text{\tiny AC}}_{\text{\tiny ATO}}\right)^2+\left(\widehat{\mbox{SE}}^{\text{\tiny BC}}_{\text{\tiny ATO}}\right)^2}.
\end{align*}

\subsection{Arbitrator Not Having Access to IPD of Covariates}

Consider the scenario where neither sponsor is willing to share their IPD of covariates with the arbitrator. Without the IPD of covariates, the arbitrator cannot run any propensity score model. Instead, the arbitrator can determine a covariate generating model (e.g., \cite{ding2025empirical}) for each sponsor to generate simulated IPD of covariates based on the AgD of their counterpart. The arbitrator also determines a propensity score model (e.g., logistic regression model) to be used by both sponsors. 

After a covariate generating model is determined and a seed is prespecified by the arbitrator, the two sponsors generate a set of covariates separately. That is, sponsor A generates the following dataset of covariates:
\begin{align*}
    \{(\bm{X}^*_i, T_i=0),\quad i=n_{\mbox{\tiny{AC}}}+1, \dots, n_{\mbox{\tiny{AC}}}+n_{\mbox{\tiny{BC}}}\}.
\end{align*}
If sponsor B publishes both the means of the variates and an estimate of the covariance matrix, then sponsor can utilize both of them in the covariate generating process. If sponsor B only publishes the means of the covariates, then sponsor A can utilize an estimate of the covariance matrix based on IPD of the trial AC to estimate that of the trial BC. Therefore, sponsor is able to run the pre-determined propensity score model based on the following data:
\begin{align*}
\{(\bm{X}_i, T_i=1),\quad i=1, \dots, n_{\mbox{\tiny{AC}}}; \quad (\bm{X}^*_i, T_i=0),\quad i=n_{\mbox{\tiny{AC}}}+1, \dots, n_{\mbox{\tiny{AC}}}+n_{\mbox{\tiny{BC}}}\}.
\end{align*}
Then sponsor A obtains the overlap weights $\omega_1$ and conduct MAIC based on the overlap weights, producing an estimate $\widehat{\theta}^{\text{\tiny AC}}_{\text{\tiny ATO}}$ and an estimate of its standard error $\widehat{\mbox{SE}}^{\text{\tiny AC}}_{\text{\tiny ATO}}$ for the estimand $\theta^{\text{\tiny AC}}_{\text{\tiny ATO}}$. 

Meanwhile, sponsor B generates the following dataset of covariates:
\begin{align*}
    \{(\bm{X}^*_i, T_i=1),\quad i=1, \dots, n_{\mbox{\tiny{AC}}}\}.
\end{align*}
Sponsor runs the pre-determined propensity score model based on the following data:
\begin{align*}
\{(\bm{X}^*_i, T_i=1),\quad i=1, \dots, n_{\mbox{\tiny{AC}}}; \quad (\bm{X}_i, T_i=0),\quad i=n_{\mbox{\tiny{AC}}}+1, \dots, n_{\mbox{\tiny{AC}}}+n_{\mbox{\tiny{BC}}}\}.
\end{align*}
Then sponsor B obtains the overlap weights $\omega_0$ and conduct MAIC based on the overlap weights, producing an estimate $\widehat{\theta}^{\text{\tiny BC}}_{\text{\tiny ATO}}$ and an estimate of its standard error $\widehat{\mbox{SE}}^{\text{\tiny BC}}_{\text{\tiny ATO}}$ for the estimand $\theta^{\text{\tiny BC}}_{\text{\tiny ATO}}$.

Finally, after receiving the results from the sponsors, the arbitrator obtains a point estimate and an estimate of its standard error for $\widehat{\theta}^{\text{\tiny AB}}_{\text{\tiny ATO}}$. 

\section{Numerical Results}

\subsection{Hypothetical Example Revisited}

The hypothetical example presented in Section 1 is revisited here. The proposed arbitrated MAIC methods are applied to estimate the ATO between drug A and drug B. 

In the hypothetical example, race is the sole effect modifier and is binary, so the IPD of race can be recovered by knowing the AgD of race (i.e., the proportions of Black and non-Black). Thus, the two proposed arbitrated MAIC methods are equivalent in this hypothetical example, and they can be described in the below three steps.

First, by the Bayesian formula, the propensity scores can be calculated as 
\begin{align*}
    \varepsilon(\mbox{Black})&=\mathbb{P}(T_i=1\mid X_i=\mbox{Black})= \frac{\mathbb{P}(X_i=\mbox{Black}\mid T_i=1)\mathbb{P}(T_i=1)}{\mathbb{P}(X_i=\mbox{Black})}\\
    &=\frac{(400/1200)(1200/2400)}{1200/2400}=\frac{1}{3},\\
    \varepsilon(\mbox{non-Black})&=\mathbb{P}(T_i=1\mid X_i=\mbox{non-Black})= \frac{\mathbb{P}(X_i=\mbox{non-Black}\mid T_i=1)\mathbb{P}(T_i=1)}{\mathbb{P}(X_i=\mbox{non-Black})}\\
    &=\frac{(800/1200)(1200/2400)}{1200/2400}=\frac{2}{3}.
\end{align*}
Therefore, the overlap weights are 
\begin{align*}
    \omega_1(\mbox{Black})&=1-\varepsilon(\mbox{Black})=\frac{2}{3}, \quad \omega_1(\mbox{non-Black})=1-\varepsilon(\mbox{non-Black})=\frac{1}{3};\\
    \omega_0(\mbox{Black})&=\varepsilon(\mbox{Black})=\frac{1}{3}, \quad \omega_0(\mbox{non-Black})=\varepsilon(\mbox{non-Black})=\frac{2}{3}.
\end{align*}

Second, sponsor A conducts MAIC based on $\omega_1$, producing an estimate of the ATO in terms of log-odds ratio comparing A and C: 
\begin{align*}
   \widehat{\theta}^{\text{\tiny AC}}_{\text{\tiny ATO}}&= \log\left[\frac{\left(180\times \omega_1(\mbox{Black})+80\times \omega_1(\mbox{non-Black})\right)\left(120\times \omega_1(\mbox{Black})+360\times \omega_1(\mbox{non-Black})\right)}{\left(20\times \omega_1(\mbox{Black})+320\times \omega_1(\mbox{non-Black})\right)\left(80\times \omega_1(\mbox{Black})+40\times \omega_1(\mbox{non-Black})\right)}\right]\\
   &= \log\left[\frac{\left(180 (2/3)+80(1/3)\right)\left(120(2/3)+360(1/3)\right)}{\left(20(2/3)+320(1/3)\right)\left(80(2/3)+40(1/3)\right)}\right]=1.30.
\end{align*}
Separately, sponsor B conducts MAIC based on $\omega_0$, producing an estimate of the ATO in terms of log-odds ratio comparing B and C: 
\begin{align*}
   \widehat{\theta}^{\text{\tiny BC}}_{\text{\tiny ATO}}&= \log\left[\frac{\left(240\times \omega_0(\mbox{Black})+100\times \omega_0(\mbox{non-Black})\right)\left(240\times \omega_0(\mbox{Black})+180\times \omega_0(\mbox{non-Black})\right)}{\left(160\times \omega_0(\mbox{Black})+100\times \omega_0(\mbox{non-Black})\right)\left(160\times \omega_0(\mbox{Black})+20\times \omega_0(\mbox{non-Black})\right)}\right]\\
   &= \log\left[\frac{\left(240 (1/3)+100(2/3)\right)\left(240(1/3)+180(2/3)\right)}{\left(160(1/3)+100(2/3)\right)\left(160(1/3)+20(2/3)\right)}\right]=1.30.
\end{align*}

Third, after receiving the results from the sponsors, the arbitrator obtains the following estimate of the ATO in terms of log-odds ratio comparing A and B:
\begin{align*}
    \widehat{\theta}^{\text{\tiny AB}}_{\text{\tiny ATO}}&=\widehat{\theta}^{\text{\tiny AC}}_{\text{\tiny ATO}}-\widehat{\theta}^{\text{\tiny BC}}_{\text{\tiny ATO}}=1.30-1.30=0.
\end{align*}

\subsection{Simulation Study}

We simulated two randomized trials: (i) trial AC comparing $A$ vs.~$C$ and (ii) trial BC comparing $B$ vs.~$C$. Within each trial, treatment assignment $Z\in\{0,1\}$ was randomized with
$\mathbb{P}(Z=1)=0.5$, where $Z=1$ denotes active treatment ($A$ in AC, $B$ in BC) and
$Z=0$ denotes control $C$.

Each subject had $\bm{X}=(X_1,X_2,X_3)^\top$ where $X_1$ is binary and $X_2, X_3$ are
continuous:
\[
X_1 \sim \mathrm{Bernoulli}(p_1),\qquad
X_2 \sim \mathcal{N}(\mu_2, 1),\qquad
X_3 \sim \mathcal{N}(\mu_3, 1).
\]
Trial-specific covariate distributions were intentionally different:
\[
(p_1^{\text{\tiny AC}}, \mu_2^{\text{\tiny AC}}, \mu_3^{\text{\tiny AC}})=(0.2, 0, 0),\qquad
(p_1^{\text{\tiny BC}}, \mu_2^{\text{\tiny BC}}, \mu_3^{\text{\tiny BC}})=(0.8, 1, -1).
\]

Continuous outcomes were generated from a linear model with effect modification:
\[
Y = \bm{X}^\top\boldsymbol{\beta} + Z(\tau_{\text{\tiny T}} + \bm{X}^\top \boldsymbol{\gamma}_{\text{\tiny T}}) + \epsilon,\qquad
\epsilon\sim\mathcal{N}(0,1),
\]
where the subscript $\text{T}=\text{A}$ or $\text{B}$, $\boldsymbol{\beta}=(0.6,-0.2,0.3)^\top$, $\tau_{\text{\tiny A}}=\tau_{\text{\tiny B}}=1$, and
\[
\boldsymbol{\gamma}_{\text{\tiny A}} = (-1, 1.5, -1)^\top,\qquad \boldsymbol{\gamma}_{\text{\tiny B}}=(1, 1.5, 1)^\top.
\]
Under this model, the marginal effect of A vs.~C in a target population with
covariate mean $\boldsymbol{\mu}_X$ is $\theta^{\text{\tiny AC}}(\mu_X)=\tau_{\text{\tiny A}}+\boldsymbol{\mu}_X^\top \boldsymbol{\gamma}_{\text{\tiny A}}$, while the marginal effect of B vs.~C in that target population is $\theta^{\text{\tiny BC}}(\mu_X)=\tau_{\text{\tiny B}}+\boldsymbol{\mu}_X^\top \boldsymbol{\gamma}_{\text{\tiny B}}$. Hence
the effect of A vs.~B in that target population is
\[
\theta^{\text{\tiny AB}}(\mu_X)
=
(\tau_{\text{\tiny A}}+\boldsymbol{\mu}_X^\top \boldsymbol{\gamma}_{\text{\tiny A}})-(\tau_{\text{\tiny B}}+\boldsymbol{\mu}_X^\top \boldsymbol{\gamma}_{\text{\tiny B}}).
\]

To calculate the overlap weights, we fit a logistic regression for trial membership $T\in\{1, 0\}$ (AC vs.~BC) to
obtain $\varepsilon(\bm{X})=\mathbb{P}(T=1\mid \bm{X})$. Overlap weights were
$\omega_{1}(\bm{x})=1-\varepsilon(\bm{x})$ for the AC trial and $\omega_{0}(\bm{x})=\varepsilon(\bm{x})$ for the BC trial, leading to the proposed methods of estimating the ATO for the overlap population with density
proportional to $\varphi(\bm{x})\varepsilon(\bm{x})(1-\varepsilon(\bm{x}))$.

Effective sample size (ESS) was computed as
\[
\mathrm{ESS}=\frac{\left(\sum_i w_i\right)^2}{\sum_i w_i^2}, 
\]
where $(w_1, \dots, w_n)$ are the weights being used, be it the MAIC weights or overlap weights. 

We also calculated the ``paradox'' rate
$I\{\widehat{\theta}^{\text{\tiny MAIC1}}_{A-B}>0
\ \text{and}\ 
\widehat{\theta}^{\text{\tiny MAIC2}}_{A-B}<0\}$, where $\widehat{\theta}^{\text{\tiny MAIC1}}_{A-B}$ and $\widehat{\theta}^{\text{\tiny MAIC2}}_{A-B}$ denote the estimated effects of $A$ vs. $B$ from the first MAIC (IPD of AC trial and AgD of BC trial) and the second MAIC (IPD of BC trial and AgD of AC trial), respectively.

We set $n_{\text{\tiny AC}}=n_{\text{\tiny BC}}=600$, repeated the above data-generating process by $1000$ times, and summarized the result in Table \ref{Tab:sim1}. In MAIC1, the target population is BC population, while in MAIC2, the target population is AC population.

Assume that larger outcome values indicate better performance. The mean estimates clearly demonstrate the MAIC paradox: MAIC1 suggests that treatment A is more effective than treatment B, whereas MAIC2 suggests the opposite conclusion. This apparent contradiction does not reflect a failure of the MAIC methodology. Rather, the two MAIC estimators target different populations and are each consistent for their respective target estimands. Importantly, the proposed estimator based on overlap weights (OW) is consistent for the ATO, thereby providing inference with respect to a common, well-defined target population.

\begin{table}[ht]\caption{Results from three estimators for three estimands}\label{Tab:sim1}
\medskip
\centering
\centering
\begin{tabular}{lrrrr}
\toprule
 & Mean Est. & True Estimand & ESE & RMSE \\
\midrule
MAIC1 (BC population) 
    & 0.358 & 0.394 & 0.430 & 0.432 \\
MAIC2 (AC population) 
    & $-0.420$ & $-0.407$ & 0.485 & 0.485 \\
OW (Overlap population) 
    & $-0.006$ & $-0.013$ & 0.208 & 0.208 \\
\bottomrule
\multicolumn{5}{l}{ESE: empirical standard error; RMSE: root mean square error}
\end{tabular}
\end{table}

We further examined the mean ESS for the three approaches. 
The average ESS was 49.8 for MAIC1 and 61.9 for MAIC2, indicating substantial loss 
of effective information relative to the nominal sample size of 600 per trial.
In contrast, the proposed OW approach retained considerably larger ESS, with mean ESS equal to 272.5 for the AC trial under weights 
$\omega_1(\cdot)$ and 272.7 for the BC trial under weights $\omega_0(\cdot)$. 
These results demonstrate that MAIC reweighting can induce severe information 
loss under limited covariate overlap, whereas overlap weighting preserves 
substantially greater efficiency. Consequently, a larger ESS results in a smaller standard error. In the simulation study, the standard error for the OW estimates was substantially lower than that for MAIC1 and MAIC2. Finally, the paradox rate was estimated at 
0.648 across 1000 Monte Carlo repetitions, indicating that the MAIC paradox 
occurred in the majority of repetitions.

\subsection{Practical Considerations}

In the simulation study, we defined the estimand using a continuous outcome variable, summarizing the population-level effect with the mean. Accordingly, the three estimators compared (MAIC1, MAIC2, and OW) were each based on the weighted sample mean.

However, the sample mean, as an estimator, does not account for covariates that capture the heterogeneity of trial populations. To address this heterogeneity and obtain more accurate estimates of treatment effects, weighted regression methods or advanced causal inference methods or doubly robust approaches \citep{hernan2020causal} can be employed.

For binary outcomes, the estimand may be expressed as a proportion difference, relative risk, or odds ratio. However, because odds ratios are non-collapsible, we recommend focusing on estimands defined by proportion difference or relative risk. For time-to-event outcomes, the estimand can be defined as a hazard ratio, survival rate difference, or restricted mean survival time (RMST) difference. However, since hazard ratios are non-collapsible, we recommend focusing on estimands based on survival rate differences or RMST difference.

In addition, MAIC is commonly used for pairwise ITC involving one internal trial and one external trial. When multiple phase 3 trials are available, such as two pivotal studies, separate MAIC analyses may be required for each trial, with additional analyses potentially conducted using integrated summary of efficacy (ISE) data. The specific approach should be determined by the pre-specified ITC protocol, which outlines all planned indirect comparison analyses.

A key limitation of MAIC is the reduction in ESS, which leads to increased variation. In our simulation study, we showed that the proposed methods using overlap weights can help mitigate ESS reduction and variance inflation to some extent. One advantage of the proposed arbitrated methods is that, if the arbitrator has access to individual patient-level covariate data from both trials, they can directly assess the degree of overlap between the two trial populations.

\section{Discussion}

One UK NICE technical support document \citep{phillippo2016nice} states that ``While there is a clear rationale for considering population-adjusted estimates of treatment effects, there is a lack of clarity about exactly how, and when, they should be applied in practice, and even whether the results are relevant to the decision problem. This increases the risk that assumptions being made in one submission are fundamentally different from---even incompatible with---the assumptions being made a year later in another on the same condition." Therefore, it is important to make sure the assumptions and estimands are compatible between different submissions. 

In this paper, we investigate the phenomenon known as the MAIC paradox. We identify its root cause and propose two arbitrated MAIC methods to resolve the paradox by estimating the average treatment effect for the overlap population.

The proposed methods are presented in the context of anchored MAIC but can be readily extended to unanchored MAIC. This extension is achieved by replacing the AC trial with a single-arm trial A, and the BC trial with a single-arm trial B, respectively.

Furthermore, we posit that a similar paradox may also arise in the context of STC. Accordingly, we refer to this general phenomenon as the PAIC paradox: 
\begin{quote} ``When there are imbalances in effect modifiers with different magnitudes of modification across treatments, contradictory conclusions may arise if PAIC (either MAIC or STC) is performed with the IPD and AgD swapped between trials.''
\end{quote}
And the root cause of the PAIC paradox is attributed to the following:
\begin{itemize}
    \item When PAIC is conducted by sponsor A, using IPD from the AC trial and aggregate data (AgD) from the BC trial, the target population is defined by the BC trial;
    \item Conversely, when PAIC is conducted by sponsor B, using IPD from the BC trial and AgD from the AC trial, the target population is defined by the AC trial.
\end{itemize}

In light of this, and by applying the same arbitration principle, corresponding arbitrated STC methods can be developed. These methods are designed to handle both cases: when the arbitrator has access to covariate IPD and when such access is not available.

To summarize, this paper offers at least two contributions. First, it proposes a new approach that HTA bodies may use as part of their review process for HTA submissions. When discrepancies are identified between two sponsors, the HTA body may serve as an arbitrator to address and resolve them. Second, it advocates the use of the average treatment effect in the overlap population (ATO) as the target estimand to mitigate the impact of limited overlap between two populations. 

Finally, an additional contribution is its potential to promote a consistent approach to handling intercurrent events (ICEs). According to ICH E9(R1), \emph{Addendum on Estimands and Sensitivity Analysis in Clinical Trials to the Guideline on Statistical Principles for Clinical Trials}, the handling of ICEs is a key attribute of the estimand definition. In conducting an arbitrated ITC, the arbitrator should ensure that both sponsors address the same ICEs and specify the same strategies for handling them.

\section*{Appendix: Theoretical Justification}

\subsection*{AC Trial}

Assumptions for the AC trial: 
\begin{enumerate}
    \item Consistency: $Y=I(Z=\text{A})Y^{A}+I(Z=\text{C})Y^C$, where $Y^Z$ is the potential outcome had the patient been treated by $Z$.
    \item Exchangeablity: $Z$ and $(Y^A, Y^C)$ are independent given $\bm{X}$. This condition is satisfied when either simple randomization is used, or when stratified randomization is implemented with $\bm{X}$ encompassing the relevant stratification factors.
    \item Positivity: $0<\mathbb{P}(Z=A\mid \bm{X}=\bm{x})=1-\mathbb{P}(Z=C\mid \bm{X}=\bm{x})<1$ for any $\bm{x}$ such that $\mathbb{P}(\bm{X}=\bm{x})>0$. This condition is also satisfied in RCTs. 
    \item Non-zero normalizing constant: $\text{Constant}=\int \varphi_1(\bm{x})\omega_1(\bm{x})\mu(d\bm{x})>0$ and $\omega_1(\bm{x})$ is a given weight function and $\varphi_1(\bm{x})$ is the marginal density of $\bm{X}$ in the AC trial. 
\end{enumerate}

Under these assumptions, the following estimand is well defined: 
\begin{align*}
    \theta^{\text{\tiny AC}} (\omega_1) &= 
    \frac{
        \int \left[ \mathbb{E}(Y \mid Z = \text{A}, \bm{X} = \bm{x}) - \mathbb{E}(Y \mid Z = \text{C}, \bm{X} = \bm{x}) \right] \varphi_1(\bm{x}) \omega_1(\bm{x})\mu(d\bm{x})
    }{
        \int \varphi_1(\bm{x}) \omega_1(\bm{x})\mu(d\bm{x}).
    } 
\end{align*}

Note that, in the above estimand, the weight function $\omega_1(\bm{x})$ serves the same purpose as $h(\bm{x})$ in \cite{li2018balancing}. Consequently, as demonstrated in \cite{li2018balancing}, the standard weighted estimators employing $\omega_1(\bm{x})$ are consistent and asymptotically normal for estimating the above estimand. We denote such estimators by $\widehat{\theta}^{\text{\tiny AC}}(\omega_1)$.

\subsection*{BC Trial}

Assumptions for the BC trial: 
\begin{enumerate}
    \item Consistency: $Y=I(Z=\text{B})Y^{B}+I(Z=\text{C})Y^C$, where $Y^Z$ is the potential outcome had the patient been treated by $Z$.
    \item Exchangeablity: $Z$ and $(Y^B, Y^C)$ are independent given $\bm{X}$. This condition is satisfied when either simple randomization is used, or when stratified randomization is implemented with $\bm{X}$ encompassing the relevant stratification factors.
    \item Positivity: $0<\mathbb{P}(Z=B\mid \bm{X}=\bm{x})=1-\mathbb{P}(Z=C\mid \bm{X}=\bm{x})<1$ for any $\bm{x}$ such that $\mathbb{P}(\bm{X}=\bm{x})>0$. This condition is also satisfied in RCTs. 
    \item Non-zero normalizing constant: $\text{Constant}=\int \varphi_0(\bm{x})\omega_0(\bm{x})\mu(d\bm{x})>0$ and $\omega_0(\bm{x})$ is a given weight function and $\varphi_0(\bm{x})$ is the marginal density of $\bm{X}$ in the BC trial. 
\end{enumerate}

Under these assumptions, the following estimand is well defined: 
\begin{align*}
    \theta^{\text{\tiny BC}} (\omega_0) &= 
    \frac{
        \int \left[ \mathbb{E}(Y \mid Z = \text{B}, \bm{X} = \bm{x}) - \mathbb{E}(Y \mid Z = \text{C}, \bm{X} = \bm{x}) \right] \varphi_0(\bm{x}) \omega_0(\bm{x})\mu(d\bm{x})
    }{
        \int \varphi_0(\bm{x}) \omega_0(\bm{x})\mu(d\bm{x}).
    } 
\end{align*}

Note that, in the above estimand, the weight function $\omega_0(\bm{x})$ serves the same purpose as $h(\bm{x})$ in \cite{li2018balancing}. Consequently, as demonstrated in \cite{li2018balancing}, the standard weighted estimators employing $\omega_0(\bm{x})$ are consistent and asymptotically normal for estimating the above estimand. We denote such estimators by $\widehat{\theta}^{\text{\tiny BC}}(\omega_0)$.

\subsection*{Arbitrated Analysis}

In particular, let $\omega_1(\bm{x})=1-\varepsilon(\bm{x})$ for the AC trial and $\omega_0(\bm{x})=\varepsilon(\bm{x})$ for the BC trial. Then the estimand in which the arbitrator is interested is 
\begin{align*}
    \theta^{\text{\tiny AB}}_{\text{\tiny ATO}} = 
    \theta^{\text{\tiny AC}}(\omega_1) - 
    \theta^{\text{\tiny BC}}(\omega_0).
\end{align*}

One additional assumption is required. For anchored ITC, the assumption of constancy of relative effects is needed, whereas for unanchored ITC, the assumption of constancy of absolute effects is needed \citep{phillippo2016nice}.

Previously, we have shown that we can construct consistent and asymptotically normal estimators $\widehat{\theta}^{\text{\tiny AC}}(\omega_1)$ and $\widehat{\theta}^{\text{\tiny BC}}(\omega_0)$ for $\theta^{\text{\tiny AC}} (\omega_1)$ and $\theta^{\text{\tiny BC}} (\omega_0)$, respectively. Therefore, the difference between the two estimators, that is, 
\begin{align*}
    \widehat{\theta}^{\text{\tiny AB}}_{\text{\tiny ATO}} = 
\widehat{\theta}^{\text{\tiny AC}}(\omega_1)- \widehat{\theta}^{\text{\tiny BC}}(\omega_0),
\end{align*}
is a consistent and asymptotically normal estimator for $\widehat{\theta}^{\text{\tiny AB}}_{\text{\tiny ATO}}$.

{\it Remark 1}: Practical issues may arise that compromise the assumptions stated above, such as (1) the presence of unmeasured effect modifiers and (2) incorrect classification of covariates as effect modifiers. To address these challenges, it is advisable to conduct sensitivity analyses to assess the robustness of the results when different sets of covariates are adjusted.

{\it Remark 2}: Once each sponsor has received the overlap weights from the arbitrator, they may proceed to analyze their respective data using these weights. While it is preferable for both sponsors to apply the same statistical method for their weighted analyses (for instance, both using PROC MEANS, PROC FREQ, or PROC GENMOD in SAS with a weight statement), this is not strictly necessary. Sponsors can use different analytical approaches, as long as each method is consistent for the trial-specific estimand. This flexibility is supported by the continuous mapping theorem: if each estimator is consistent for its corresponding estimand, then the difference between these estimators will likewise be consistent for the difference between the estimands.



\bibliographystyle{apalike}
\bibliography{references}
\end{document}